\documentclass{article} 
\usepackage{iclr2025_conference,times}

\usepackage{amsmath,amsfonts,bm}

\def\eqref#1{equation~\ref{#1}}

\def\1{\bm{1}}

\DeclareMathAlphabet{\mathsfit}{\encodingdefault}{\sfdefault}{m}{sl}
\SetMathAlphabet{\mathsfit}{bold}{\encodingdefault}{\sfdefault}{bx}{n}

\usepackage{hyperref}
\usepackage{url}

\usepackage{amsmath,amssymb,amsfonts}
\usepackage{algorithmic}
\usepackage{graphicx}
\usepackage{textcomp}
\usepackage{xcolor}

\usepackage{mathtools}
\usepackage{amsthm}
\usepackage{multirow}
\usepackage{booktabs}
\usepackage{epigraph} 

\newcommand{\ModelName}{{LaMsS}}
\newcommand{\vanillaFT}{{VanillaFT}}

\title{LaMsS: When Large Language Models Meet Self-Skepticism}

\author{
Yetao Wu\textsuperscript{1}\thanks{Equal contribution.}, 
Yihong Wang\textsuperscript{1}\footnotemark[1],
Teng Chen\textsuperscript{1},
Ningyuan Xi\textsuperscript{2}\thanks{This work was done when Ningyuan Xi was an intern at Geely.},
Qingqing Gu\textsuperscript{1},
Hongyang Lei\textsuperscript{1},
\textbf{Luo Ji\textsuperscript{1}\thanks{Corresponding author.}} \\
\textsuperscript{1}Geely AI Lab \\
\textsuperscript{2} Beihang University\\
\texttt{\{Yetao.Wu, YiHong.Wang2, Teng.Chen2, Ningyuan.Xi,Qingqing.Gu3,Hongyang.Lei,} \\
\texttt{Luo.Ji1\}@geely.com}\\
}

\iclrfinalcopy 
\begin{document}

\maketitle

\begin{abstract}
Hallucination is a major challenge for large language models (LLMs), preventing their further application in some fields. The skeptical thinking of humankind could be useful for LLMs to self-cognition, self-reflection and alleviate their hallucinations. Inspired by this consideration, we propose a novel approach called LaMsS, which combines the semantic understanding capability of LLMs with self-skepticism. By introducing a series of skepticism tokens and augmenting them into the vocabulary, we conduct both pertaining and finetuning, which allow the LLM to decode each normal token followed by a skeptical token, representing different skepticism levels. By calculating the response skepticism given a query, one can define a new self-aware LLM which is only willing to answer with relative lower skepticism level than the threshold. By examining the accuracy, AUC and AP of willingly answering questions, we demonstrate that LaMsS achieves better performance than baselines on both multi-choice questions and open-domain question-answering benchmarks, and can generalize to multi-task and out-of-domain settings. Our study sheds some lights on the self-skepticism modeling on further artificial intelligence. Project code and model checkpoints can be found in \url{https://anonymous.4open.science/r/SM-1E76}.
\end{abstract}

\section{Introduction}
\label{sec:intro}




Large Language Models (LLMs) have revolutionized natural language processing and artificial intelligence, demonstrating remarkable task-agnostic capabilities across a wide range of fields \citep{naveed2024comprehensiveoverviewlargelanguage,zhao2023surveylargelanguagemodels, minaee2024largelanguagemodelssurvey, openai2024gpt4technicalreport, touvron2023llamaopenefficientfoundation}. Despite these remarkable achievements, the generative nature of LLM simultaneously raises the challenge of hallucination \citep{huang2023surveyhallucinationlargelanguage,bai2024hallucinationmultimodallargelanguage}. The hallucination issues are twofold: i) upon a knowledge-related question, an instruction fine-tuned LLM might provide a plausible yet fabricated, mistaken answer; ii) as a pretrained LLM, the tendency to continue a prompted text (maybe hallucinated itself) with although fluent but factually incorrect texts. Such phenomena affect LLM's trustworthiness and prevent further widespread application of LLMs especially for high-level expertise fields, such as healthcare, legal, finance, and manufacture  \citep{Ji2023Survey,Zhang2023Siren's}.

\begin{figure}[t]
\centering
\includegraphics[width=0.98\linewidth]{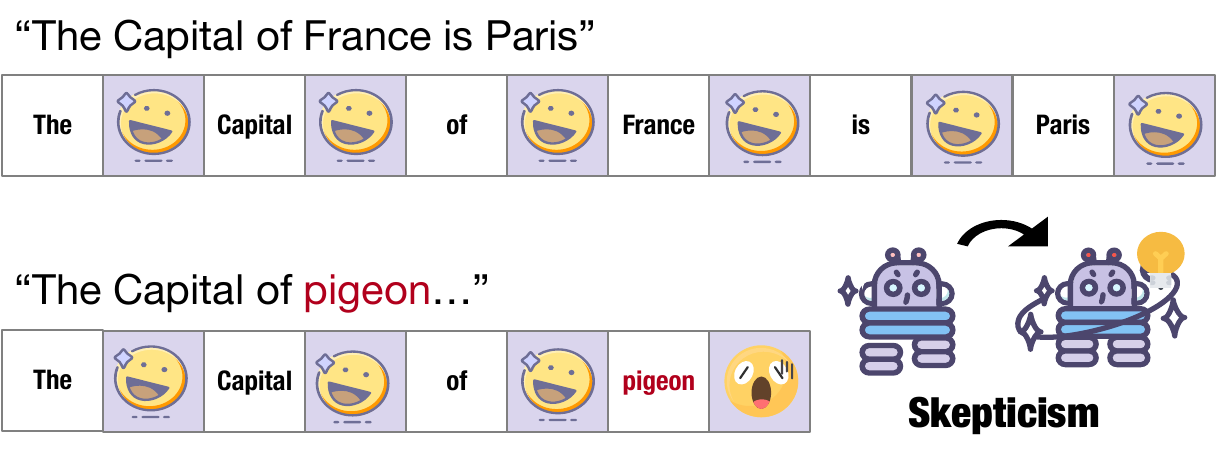}
\caption{Paradigm of Self-Skepticism by LLM. \\
The emojis represent the self-skepticism levels of the `formal' tokens by LLM itself. Problematic, counterfactual phrases (\textit{e.g.}, `pigeon' after `capital') arouse suspicious and skeptical feelings.}
\label{paradigm}
\end{figure}

Different aspects of studies have been proposed to alleviate the problem of hallucinations in LLM, including utilizing the model inherent log probabilities \citep{Lin2022Teaching,kadavath2022languagemodelsmostlyknow,huang2023survey}, augmenting uncertainty tokens \citep{Zhang2024R-Tuning}, using external knowledge \citep{peng2023checkfactstryagain} or an extra examiner agent \citep{cohen-etal-2023-lm}. However, most of these works focus on mitigating hallucinations within the model response, while generally neglecting the doubt checking and uncertainty assessment on the prompt or user query. Due to the causal inference mechanism on the decoder-only model, a fabricated prefix text would mislead the LLM, generating problematic texts. Considering these situations, we argue that a `doubtful' LLM which consistently assesses the plausibility of all textual tokens (not only its own response) might have a deeper insight and generate more factual correct responses in higher quality.



Our work is motivated by the theory of the famous philosopher Ren\'e Descartes, in which skepticism, or `hyperbolic doubt', plays a crucial role in rationalized thinking and ultimately helps find out the unquestionable facts \citep{descartes1641meditations}. The emotion-as-information theory \citep{schwarz1983mood} suggests that the skeptical feeling can lead to a more careful information examination; later studies further show that self-skepticism is a core component of critical thinking \citep{facione1990critical} and meta-cognitive experiences \citep{koriat1999information}, and can be viewed as the first principle of rationality, as articulated by Richard Feynman \citep{Feynman1974notfool}. For a simple instance, when questioned with "What is the \textcolor{red}{capital} of \textcolor{red}{pigeon}?" or "How many \textcolor{red}{eyes} does the \textcolor{red}{finger} have?", humans are inclined to have skeptical emotions and argue with the reasonability of the question itself, instead of forced answering. Furthermore, given an accurate but challenging question such as a mathematical test, one might doubt himself and refuse to answer with low confidence. Inspired by these theoretic observations, we speculate concurrently training with both semantic knowledge and self-skepticism capabilities would produce more inherent knowledgeable and self-consistent language models (Figure \ref{paradigm}).







\epigraph{Well, then, what am I? A thing that thinks. What is that? A thing that \textbf{doubts}, understands, affirms, denies, wants, refuses, and also imagines and senses. [Meditations on First Philosophy: 2nd Meditation Part2]}{\textit{Ren\'e Descartes}}

\begin{figure*}[t]
    \centering
    \includegraphics[width=1\linewidth]{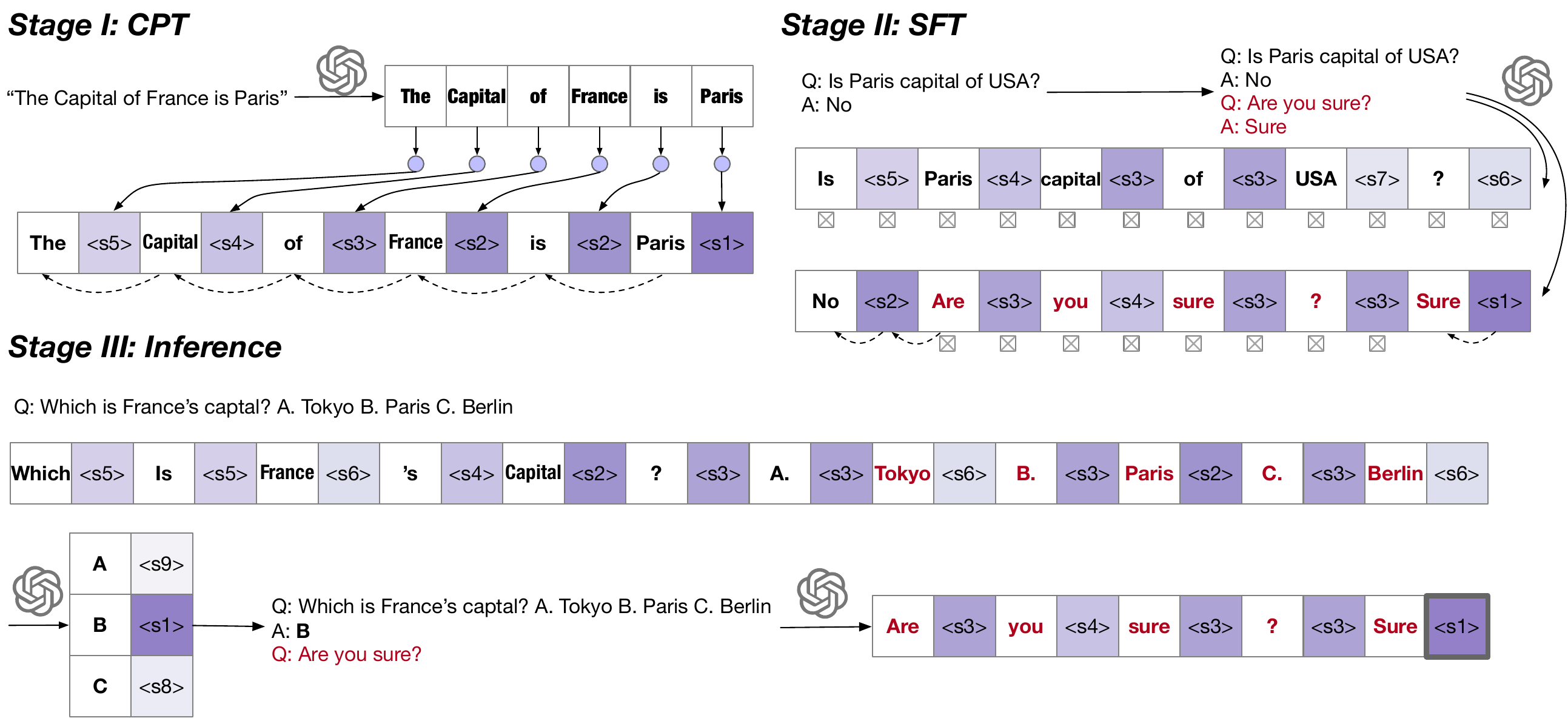}
    \caption{Detailed Framework of \ModelName.\\
    Stage I: first learn the plausibility of tokens from pretrained LLM, then continual pretraining on the corpus with vocabulary augmented with skepticism tokens.\\
    Stage II: augment the QA pair with the question 'Are you sure/unsure', inference the continual pretrained LLM to answer this augmented question, and finally finetune on these two QA pairs.\\
    Stage III: first inference on the finetuned LLM, get the most plausible answer, then concatenate with the augmented question, and inference the second time to obtain the skepticism probability.}
    \label{fig:framework}
\end{figure*}

In this paper, we propose \textbf{{\ModelName}}, an innovative paradigm to augment \textbf{La}nguage \textbf{M}odel with \textbf{s}elf-\textbf{S}kepticism thinking ability. To help LLM self-skepticism, we define a series of `skepticism tokens' to discretely represent the skeptical levels, with the tokenization vocabulary expanded. The tokenized text is then reformulated as a sequence with each original text token (we call it the `formal' token) followed by a skepticism token. We then pretrain the LLM with substantial text corpus, where we self-regress the skepticism token from the preceding tokens, as an auxiliary loss to the conventional self-regression on normal tokens. By such a paradigm, we let the LLM learns the plausibility from plausible texts, such that LLM becomes proficient on skepticism tokens, with self-awareness of suitable skepticism grounded by preceding contents. The query-response finetuning stage then follows, with an extra rethinking question augmented similar to R-tuning \citep{Zhang2024R-Tuning}, to further strengthen the skeptical feeling accuracy by the annotated ground truths. During the inference stage, the model sequentially decodes normal tokens and corresponding skepticism tokens, providing more plausible responses and self-assessments. Figure \ref{fig:framework} exhibits the framework of {\ModelName}. In summary, our contributions are:

\begin{itemize}
    \item We design a new paradigm to feed LLM the skepticism thinking, similar to humankind. Self-skepticism is operated by skeptical tokens, learning by both pertaining and finetuning stages.
    \item Our method not only generates more reasonable answers but also self-estimates the plausibility of the prompt or query text, which is often neglected by hallucination-related studies.
    \item We conduct rigorous experiments to verify that {\ModelName} achieves state-of-the-art (SOTA) performance on both multi-choice and open-domain QA benchmarks, and can generalize from in-domain to out-of-domain test sets.
\end{itemize}

\section{Method}
\label{method}

In this section, we first introduce our {\ModelName} method, which integrates skeptical tokens into the vocabulary and includes three stages:  continual pre-training, supervised fine-tuning and inference. The entire framework of {\ModelName} is visualized in Figure \ref{fig:framework}.

\subsection{Tokenization and Annotation of Skepticism}

Our methodology starts by defining the skepticism tokens to represent the model's self-skepticism levels, in a discrete manner. In more detail,  we augment the tokenizer vocabulary with special tokens $s \in \{ <s_0>, <s_1>, \dots, <s_9> \}$, with the level of skepticism higher as the subscript increases. 

Given a tokenized text sequence $[z_0, z_1, \dots, z_L]$, with $L$ is the total sequence length. For each position index $i \in [0, 1, \cdots, L]$, we append a skepticism token $s_i$ to the right of each normal token $z_i$, such that the token sequence becomes:
\begin{equation}
[(z_0, s_0), (z_1, s_1), \dots, (z_L, s_L)] \label{eq:token_seq}
\end{equation}

We quantitatively correlate the skepticism tokenization with the log scale of log probability of precedent norm tokens. By performing a forward pass of raw text corpus from a pretrained LLM, we obtain the log probability (denoted by $\log \left[\text{Prob}(\cdot) \right]$) of each normal token, which is then recorded and discretized to convert into the ground truth skepticism token:

\begin{equation}
\hat{s}_i \leftarrow "<s_k>", \quad \text{if} \label{eq:skepticism_level} -\log \left[\text{Prob}(z_{i}|\mathbf{z}_{0:i-1}) \right] \in [ k, k+1 ), \quad k = 0, 1, \dots, 9 \notag
\end{equation}
in which we use the abbreviation expression $\mathbf{z}_{0:t} := z_{1...t}$. Note for the extremely skeptical cases, $\hat{s}_i$ is also annotated with "$<s_9>$", \textit{i.e.}, when $-\log \left[\text{Prob}(z_{i}|\mathbf{z}_{0:i-1}) \right] \leq 10$.

\subsection{Stage I: Continual Pre-Training}


Training of {\ModelName} starts from a pretrained checkpoint of LLM, with $\theta$ as its learnable parameter. Given the new token sequence (Eq (\ref{eq:token_seq})), we conduct Continual Pre-Training (CPT) on it with the CPT loss expressed as


\begin{align}
    \mathcal{L}^{PT}_i &= - \log \left[\text{Prob}(z_{i}|\mathbf{z}_{0:i-1}, \theta) \right] \label{eq:pt_loss} \\
    \mathcal{L}^{S}_i &= - \log \left[\text{Prob}(s_{i}|\mathbf{z}_{0:i}, \theta) \right] \label{eq:skeptism_loss} \\
    \mathcal{L}^{CPT} &= \frac{1}{L} \sum_{i=0}^{L} \mathcal{L}^{PT}_i + \mathcal{L}^{S}_i \label{eq:cpt_loss}
\end{align}
where $\mathcal{L}^{PT}$ is the cross-entropy loss for conventional pertaining, while $\mathcal{L}^{S}$ is the loss for skepticism tokens pertaining. By augmenting $\mathcal{L}^{S}$ to $\mathcal{L}^{PT}$ during CPT, we post-adapt the LLM to the new paradigm where normal tokens and skepticism tokens are always paired.




\subsection{Stage II: Supervised Finetuning}

During the Supervised finetuning(SFT) stage, we create our skepticism-aware data by two-pass. First, given a user query, we inference our CPT-version model to generate the original response and corresponding log probabilities (Equation (\ref{eq:ori_qa})); second, we augment that QA pair with an extra prompt, and generate its answer "Sure/Unsure." (Equation (\ref{eq:aug_qa})):
\begin{align}
    &\text{\{Query\}} \xRightarrow{\text{LLM}} \text{\{Response\}}, \text{Probs} \label{eq:ori_qa} \\
    &\text{\{Query\}} \text{\{Response\}} \text{\{aug prompt\}}  \xrightarrow{\text{Probs} > \epsilon ?} \text{\{answer\}} \label{eq:aug_qa}
\end{align}
Similar to R-tuning \citep{Zhang2024R-Tuning}, $\text{\{aug prompt\}}$ is implemented with "Are you sure you accurately answered the question based on your inherent knowledge?". Depending on a predefined probabilistic threshold $\epsilon$, $\text{\{answer\}}$ is "Sure" if the probability is larger than $\epsilon$; and "Unsure" if smaller than $\epsilon$. Besides the normal tokens, we finally determine the self-skepticism token from log probability results by Equation (\ref{eq:skepticism_level}).  



We then conduct SFT by viewing normal and skepticism tokens as a uniform sequence, as denoted by Equation (\ref{eq:token_seq}). The SFT loss is 
\begin{equation}
    \mathcal{L}^{SFT} = - \frac{1}{L}\sum_{i=1}^{L} \log \left[ \text{Prob}(y_{i+1}|\mathbf{x}, \mathbf{y}_{0:i}, \theta) \right]
    \label{eq:sft_loss}
\end{equation}
where $\mathbf{x}$ is the token union of \{Query\} and \{aug prompt\}, $\mathbf{y}$ is the token union of \{Response\} and \{aug answer\} in Equation (\ref{eq:aug_qa}), and $L$ is the length of $\mathbf{y}$.




\subsection{Stage III: Inference}

Given a user query, the model trained by Stage II is employed to generate the response, decoding each normal token and its skepticism token sequentially. After that, we again augment with the prompt "Are you sure you accurately answered the question based on your inherent knowledge?", then inference the second time. The final skepticism level can be obtained from the relative weighting between the log probability of `sure' and `unsure' tokens.

Although we can obtain the skepticism levels from the log probability of skepticism tokens in the original response, here we still utilize the augmented QA pair to extract the self-skepticism level, which is based on a single token, instead of the whole token sequence. The experiment result indicates the superiority of this method.





\begin{table*}[h!]
\caption{Details of Training Datasets. MCQ means multiple choice question and QA means Question-Answering. For dataset sizes, the token numbers are listed for CPT datasets and the number of samples are listed for SFT datasets.}
\label{tab:datasets}

\begin{center}
\scriptsize
\begin{tabular}{ccccc}
\toprule
\multicolumn{1}{c}{\bf Stage}  &\multicolumn{1}{c}{\bf Datasets} &\multicolumn{1}{c}{\bf  Format}  &\multicolumn{1}{c}{\bf Subsets}   &\multicolumn{1}{c}{\bf Size}   \\ 
\toprule
\multirow{ 5}{*}{CPT}    
& \multirow{ 2}{*}{Dolma~\citep{soldaini2024dolmaopencorpustrillion}}  & \multirow{ 2}{*}{raw text }  &  gutenberg books & 3.74B  \\
& & & wiki & 3.32B  \\
\cmidrule{2-5}
& \multirow{ 3}{*}{Pile~\citep{Gao2020The}}  & \multirow{ 3}{*}{raw text }  &  opensubtitle & 0.10B  \\
& & & arxiv abstract & 0.83B   \\
& & & pubmed abstract & 0.21B   \\
\midrule
\multirow{ 10}{*}{SFT}  
& \multirow{ 2}{*}{MMLU~\citep{Hendrycks2020Measuring}}  & \multirow{ 2}{*}{MCQ}  &  ID   & 2439  \\
&  & & OOD   & 9155  \\
\cmidrule{2-5}
& \multirow{ 2}{*}{WiCE~\citep{Kamoi2023WiCE:}}  & \multirow{ 2}{*}{MCQ}  &  Train   & 3470  \\
&    & & Test  & 958  \\
\cmidrule{2-5}
& \multirow{ 2}{*}{FEVER~\citep{Thorne2018FEVER:}}  & \multirow{ 2}{*}{MCQ}  &  Train     & 9999 \\
&  & & Test      & 9999 \\
\cmidrule{2-5}
& \multirow{ 2}{*}{ParaRel~\citep{Elazar2021Measuring}}  & \multirow{ 2}{*}{QA}  &  ID   & 5584  \\
&  & & OOD   & 13974  \\
\cmidrule{2-5}
& \multirow{ 2}{*}{HotpotQA~\citep{Yang2018HOTPOTQA:}}  & \multirow{ 2}{*}{QA}  &  Train        & 10000  \\
&  & & Test     & 7405  \\
\bottomrule
\end{tabular}
\end{center}
\end{table*}

\section{Experiments}
\label{experiment}

In this section, we first introduce the training and evaluation datasets, then the experimental settings including implementation details, comparable baselines, evaluation metrics and tasks. We finally provide the formal experiment results and some typical cases.

\subsection{Datasets}

In this subsection, we brief introduce the dataset details. Table \ref{tab:datasets} summarizes statistics, format and citations of the datasets. For the `size' column, we list the token numbers for CPT datasets and the number of samples for SFT datasets.

\noindent \textbf{Pretraining Datasets:} 
Training dataset on the CPT stage is a mixture of several subsets, including Gutenberg books, wiki, opensubtitle, arxiv abstract and pubmed abstract.


\noindent \textit{Gutenberg books and wiki} are from Dolma, which is an open corpus of 3T tokens, encompassing 5 billion documents sourcing from the web, scientific literature, code, public domain books, social media, and encyclopedias. 

\noindent \textit{Opensubtitle, arxiv abstract and pubmed abstract} are from Pile, which is a 825 GB corpus of English text derived from academic or professional sources.




\noindent \textbf{Finetuning Datasets:} 
Datasets used for SFT can be classified into the following two categories:


\noindent \textit{Multiple-Choice Question (MCQ)}: including \textbf{MMLU}, \textbf{WiCE}, and \textbf{FEVER}. The question provides several options and the model needs to choose one correct option. 

\noindent \textit{Question-Answering (QA)}: including \textbf{ParaRel} and \textbf{HotpotQA} . For these two datasets, the model needs to generate an open-form answer. 

To further evaluate the model's capability on various test distributions, we keep consistent with the configuration of R-tuning \citep{Zhang2024R-Tuning}, in which \textbf{MMLU} and \textbf{ParaRel} test sets are further classified into in-domain and out-of-domain subsets. For brevity, in the following context we use ID and OOD to denote in-domain and out-of-domain, respectively. For ease of fair comparison, we download ID and OOD subsets from \citep{Zhang2024R-Tuning} directly. 





\subsection{Setting}

\noindent \textbf{Implementation:}
We choose Qwen2-7B \citep{qwen2techreport2023} as the base model in our experiments. CPT is running on 128 Nvidia A100-80GB GPUs and SFT is running on 8 GPUs. We use accelerator \footnote{\url{https://github.com/microsoft/DeepSpeed/blob/master/deepspeed/accelerator}} and deepspeed \footnote{\url{https://github.com/microsoft/DeepSpeed}} to run the experiment. Appendix shows experimental hyperparameters.



\noindent \textbf{Baselines:}
Starting from the same pretrained checkpoint, we consider the following baselines:


\noindent \textit{\vanillaFT}: the vanilla fine-tuning approach based on the same training datasets.

\noindent \textit{R-tuning \citep{Zhang2024R-Tuning}}: an instruction tuning approach which teaches LLMs to identify and refrain from answering questions beyond their parametric knowledge, thereby mitigating the hallucination issue. 

\noindent \textbf{Tasks:}
Similar to \citet{Zhang2024R-Tuning}, here we conduct two types of experiments, single-task and multi-task. The single-task experiment studies the performance training by the individual dataset, while the multi-task experiment evaluates models trained by the mixture of datasets. 



\subsection{Evaluation}

Models are measured with three metrics: accuracy (ACC), Average Precision (AP) and AUC.

\noindent \textbf{ACC:}
In this work we exhibit the willing-accuracy, that is, the fraction of correctly answered questions over the questions the model willingly answers:
\begin{equation}
    \text{ACC} = \frac{\text{\# of correctly answered questions}}{\text{\# of willing answered questions}}.
\end{equation}
To be consistent with the training setting, we use the same skeptical threshold $\epsilon = 0.5$ to judge if the model is `willing' to answer.

\noindent \textbf{AP:}
The Average Precision (AP) score provides a manner to summarize the precision-recall curve into a single representing value: 
\begin{equation}
    \text{AP} = \sum_{k=0}^{n-1} (R(k+1) - R(k)) \times P(k)
\end{equation}
where $n$ is the number of data, $k$ is the number of data selected for the current threshold, $P$ and $R$ denote precision and recall. A high-accuracy model assigns correct answers with high confidence and hallucinated answers with low confidence, leading to a high AP score.


\noindent \textbf{AUC:} AUC (Area Under the ROC Curve) is the area under the ROC (Receiver Operating Characteristic) curve, the higher the AUC value, the better the classification performance. ROC depicts the performance of the classifier at different thresholds by taking the true positive rate (TPR) and the false positive rate (FPR) as the horizontal and vertical coordinates:
\begin{equation}
    \text{TPR} = \frac{\text{TP}}{\text{TP + FN}}, \quad \text{FPR} = \frac{\text{FP}}{\text{FP + TN}}
\end{equation}
where TP (True Positive) is the number of correctly recognized positive cases, FN (False Negative) is the number of incorrectly recognized positive cases as negative cases, FP (False Positive) is the number of incorrectly identified negative examples as positive, while TN (True Negative) represents the number of correctly identified negative examples.


\begin{table}[t]
\caption{Single-task experimental results on MMLU, WiCE, Fever, Parallel and HotpotQA with AP, AUC and ACC scores (\%). MMLU and Parallel are classified into subsets of ID (in-domain) and OOD (out-of-domain), respectively.}
\label{singe-task result}

\begin{center}
\scriptsize
\begin{tabular}{c|c|c|ccc}
\toprule
Dataset & Domain & Metric & {\vanillaFT} & R-tuning & {\ModelName} \\ 
\toprule
     
\multirow{6}{*}{MMLU} 

& \multirow{3}{*}{ID} 
& AP & 37.04 & 86.89 & \textbf{88.83} \\
    &        & AUC & 56.40 & 70.91 & \textbf{73.12} \\
   &      & ACC &  68.63 & \textbf{69.37} & 69.00\\
\cmidrule{2-6}

& \multirow{3}{*}{OOD} 
 & AP & 37.71 & 85.78 & \textbf{88.18} \\
    &        & AUC & 58.94 & 68.93 & \textbf{72.33} \\
   &      & ACC & 69.10 & 68.92 & \textbf{69.11} \\
\hline
    
\multirow{3}{*}{WiCE} 

& \multirow{3}{*}{FULL} 
 & AP & 67.14 & 56.92 & \bf 85.79 \\ 
    &        & AUC & 46.88 & 50.88 & \bf 77.17 \\ 
   &      & ACC & 29.59 & 55.11 & \bf 67.35 \\ 
   
    
\hline
    
\multirow{3}{*}{Fever} 

& \multirow{3}{*}{FULL} 
 & AP & 47.59 & 90.08 & \textbf{96.99} \\
    &        & AUC & 63.28 & 73.37 & \textbf{78.55} \\
   &      & ACC & 56.75 & 73.34 & \textbf{91.64} \\
   
    
\hline
     
\multirow{6}{*}{Parallel} 

& \multirow{3}{*}{ID} 
 & AP & 59.25 & \textbf{92.16} & 86.52 \\
    &        & AUC & 26.82 &  30.92 & \bf 64.95 \\
   &      & ACC & 24.02 & 29.33 & \textbf{59.40}\\
\cmidrule{2-6}

& \multirow{3}{*}{OOD} 
  & AP & 57.08 & \textbf{87.72} & 64.95 \\
    &        & AUC & 29.56 & 17.38 & \textbf{45.97} \\
   &      & ACC & 19.31 & 11.47 & \textbf{37.96} \\
\hline
    
\multirow{3}{*}{HotpotQA} 

& \multirow{3}{*}{FULL} 
  & AP & 61.55 & \textbf{68.63} & 63.95 \\
    &        & AUC & 29.56 & 17.38 & \bf 45.97 \\
   &      & ACC & 19.31 & 11.47 & \bf 37.96 \\
\bottomrule   

\end{tabular}
\end{center}


\end{table}

\subsection{Single-task Results}

Table \ref{singe-task result} lists the results of single-task experiments. {\ModelName} surpasses \vanillaFT and R-Tuning on all MCQ benchmarks including MMLU, WiCE and Fever. It suggests that {\ModelName} enjoys a reasonable modeling of skepticism, assigns reasonable confidence to self-answers, and finally provide accurate answers when it is willing to respond.

Table \ref{singe-task result} also indicates {\ModelName} has superior performance for open-domain QA datasets like Parallel and HotpotQA, mainly on AUC and ACC. This result indicates that {\ModelName} has high robustness in skeptical thinking on different scenarios and question formats, and good generalization ability for out-of-domain texts.

\begin{table}[t]
\caption{Multi-task experimental results in percentage on MMLU, WiCE and Fever with AP, AUC and ACC scores (\%). MMLU is classified into subsets of ID (in-domain) and OOD (out-of-domain), respectively.}
\label{multitask result}

\begin{center}
\scriptsize
\begin{tabular}{c|c|c|ccc}
\toprule

Dataset & Domain & Metric & {\vanillaFT} & R-tuning & {\ModelName} \\ 

\toprule
     
\multirow{8}{*}{MMLU} 

& \multirow{3}{*}{ID} 
 & AP & 36.12 & \bf 87.45 & 87.16 \\ 
    &        & AUC & 57.37 & 73.73 & \bf 78.22 \\ 
   &      & ACC & \bf 69.13 & 66.54 & 66.83\\ 
\cmidrule{2-6}

& \multirow{3}{*}{OOD} 
  & AP & 37.99 & 86.70 & \textbf{88.04} \\ 
    &        & AUC & 59.08 & 69.66 & \bf 79.02 \\ 
   &      & ACC & \bf 69.19 & 66.84 & 67.46 \\ 
\hline
    
\multirow{3}{*}{WiCE} 

& \multirow{3}{*}{FULL} 
  & AP & 31.12 & 63.14 & \textbf{88.17} \\ 
    &        & AUC & 42.6 & 45.38 & \textbf{70.63} \\ 
   &      & ACC & 36.32 & 32.88 & \bf 71.92 \\ 

\hline
    
\multirow{3}{*}{Fever} 

& \multirow{3}{*}{FULL} 
 & AP & 59.94 & 87.43 & \textbf{93.60} \\ 
    &        & AUC & 46.41 & \textbf{77.61} & 75.05 \\ 
   &      & ACC & 35.83 & 74.38 & \textbf{81.13} \\ 

\bottomrule

\end{tabular}
\end{center}

\end{table}

\subsection{Multi-task Results}

Table \ref{multitask result} lists the MCQ results of multitask experiments, still in terms of AP, AUC and ACC scores. Similar to the single-task experiment, {\ModelName} still outperforms {\vanillaFT} and R-tuning, with seldom exceptions. This result indicates that {\ModelName} has good scaling capability and can generalize to complicated, task-mixing scenarios. By scaling up to even more datasets, tasks and domains, one can expect that {\ModelName} would align with semantic understanding better and emerge deeper skepticism thinking.

\subsection{Ablation Study}

To verify the effectiveness of each component of {\ModelName}, we implement the following ablation approaches:
\begin{itemize}
    \item no-CPT: conduct the finetuning phase directly, without the pretraining phase.
    \item no-Aug: to obtain the self-skepticism level, average the decoded skepticism tokens in the response, instead of augmenting the question as in Equation (\ref{eq:aug_qa}).
    \item no-$\epsilon$: do not use the skepticism threshold $\epsilon$, instead determine the binary confidence by comparison between self-inferenced answers and ground truth.
\end{itemize}

We conduct the ablation experiments on both ID and v subsets of MMLU, with results shown in Table \ref{ablation}. Results reveal that the complete {\ModelName} method performs better than its variants, across various metrics. These results highlight the effectiveness of the entire {\ModelName} framework.


\begin{table}[t]
\caption{Ablation study of {\ModelName} on MMLU, compared to no-Aug and no-Threshold.} %
\label{ablation}

\centering
\scriptsize
\begin{tabular}{c|c|c|cccc}
\toprule

Dataset & Domain & Metric & no-CPT & no-Aug & no-$\epsilon$ & {\ModelName} \\ 

\toprule
     
\multirow{8}{*}{MMLU} 

& \multirow{4}{*}{ID} 
 & AP & 84.86 & 66.99 & 70.24 & \textbf{88.83} \\
    &        & AUC & 67.76 & 64.21 & 67.21 & \textbf{73.12} \\
   &      & ACC & 62.44 & 56.46 & 63.84 & \textbf{69.00}\\
\cmidrule{2-7}
& \multirow{4}{*}{OOD} 
 & AP & 87.52 & 61.59 & 68.75 & \textbf{88.18} \\
    &        & AUC & 69.52 & 53.50 & 57.56 & \textbf{72.33} \\
   &      & ACC & 65.10 & 61.07 & 62.21 & \textbf{69.11} \\
\bottomrule
\end{tabular}

\end{table}

\begin{figure*}[htbp!]
    \centering
    \includegraphics[width=0.48\linewidth]{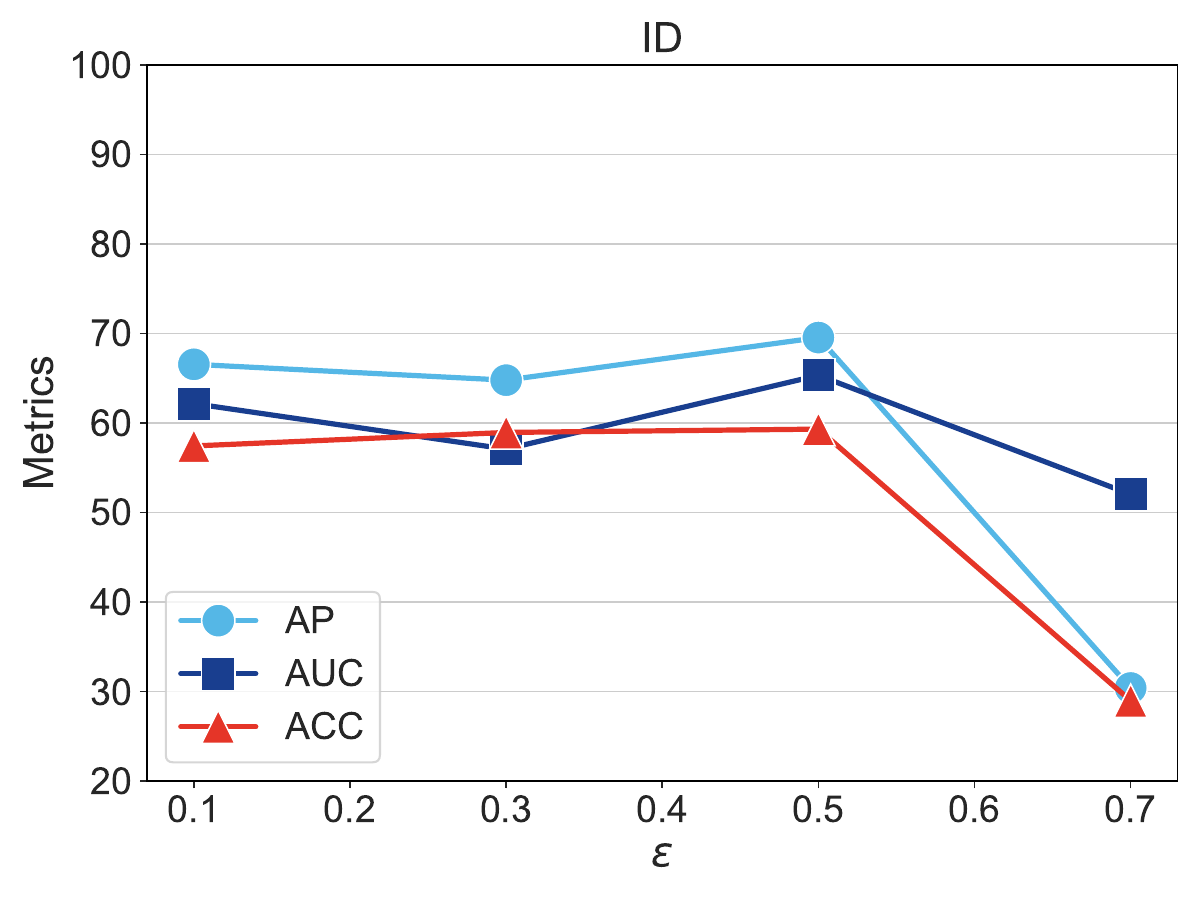}
    \hspace{0.1in}
    \includegraphics[width=0.48\linewidth]{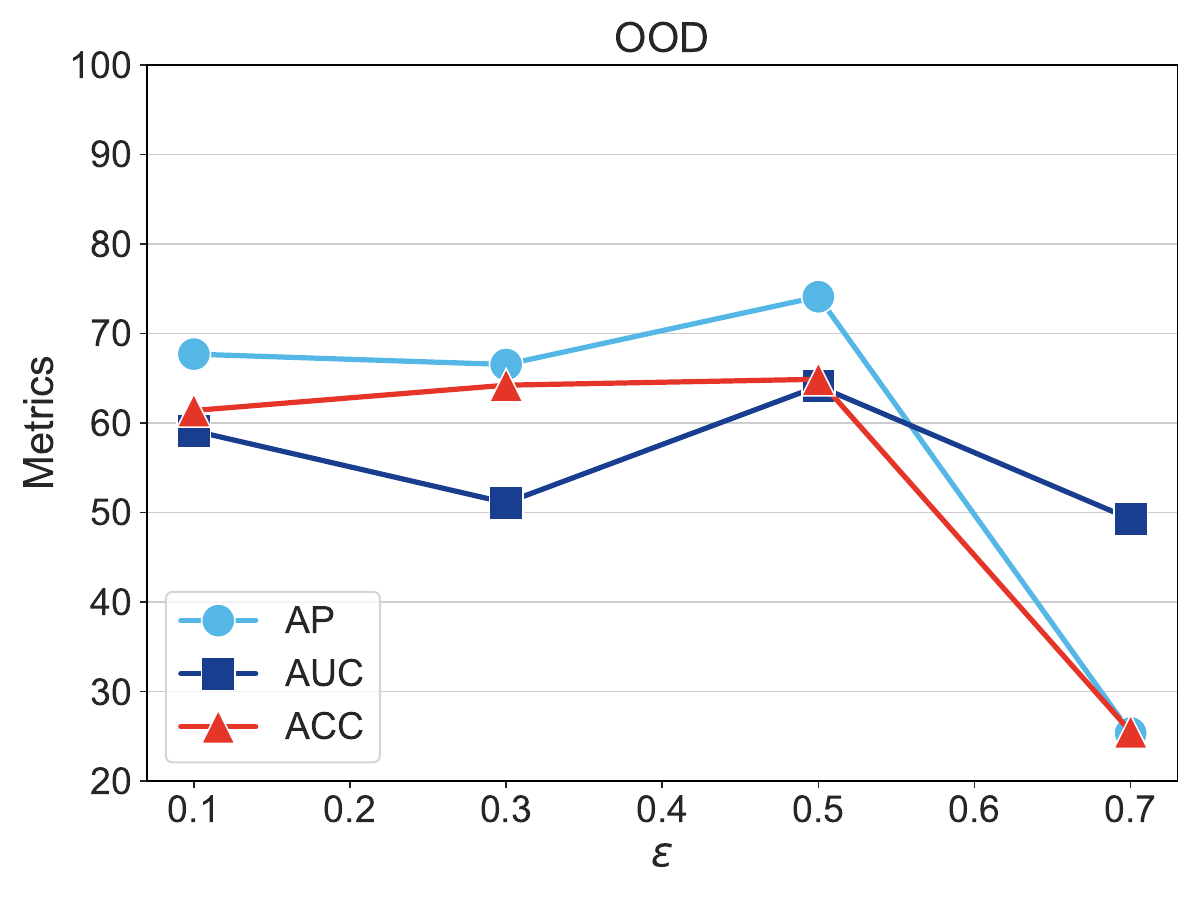}
    \caption{Sensitivity Plots of MMLU Metrics as Functions of Skepticism Thresholds $\epsilon$. Left: ID; Right: OOD.}
    \label{sensitivity}
\end{figure*}

\subsection{Sensitivity Study}

The skepticism threshold $\epsilon$ plays a critical role in our approach. To verify our choice, here we further conduct its sensitivity analysis, as indicated in Figure \ref{sensitivity}. The sensitivity plots illustrate various metrics on MMLU as functions of $\epsilon$, for both ID and OOD tests. A lower threshold may lead to more conservative predictions (higher skepticism), while a higher threshold results in more optimistic predictions (lower skepticism). The peaks of both ID and OOD curves are at 0.5, which indicates $\epsilon = 0.5$ is potentially an optimal choice for our experiments, striking an optimal balance between the skeptical sensitivity and the answer willingness.





\subsection{Typical Cases}

To better illustrate the mechanism of {\ModelName}, we highlight several typical cases. First we revisit the exampled statements proposed in Section \ref{sec:intro}:
\begin{align}
    &\text{The \textcolor{brown}{capital} of \textcolor{blue}{France}} (2.1\text{e-3}) <s_2> \notag \\
    &\text{The \textcolor{brown}{capital} of \textcolor{red}{pigeon}} (9.6\text{e-5}) <s_4>
\end{align} 
in which the number within the parentheses are token probabilities recognized by LLM, and the angle brackets denote the skepticism tokens, as introduced in Section \ref{method}. With key contextual phases in brown, the consistent phases are in blue and the inconsistent or counterfactual phases are in red. In this example, comparing with `France', our {\ModelName} assigns the counterfactual phase (\textit{i.e.}, `pigeon' after the `capital') a low probability then decode a high-level skepticism token. Similar behaviors can still be ensured by longer and more complicated expressions, for instance:
\begin{align}
    \text{If I want to visit Paris in \textcolor{brown}{autumn}, I would like go in} & \notag \\ 
    \text{\textcolor{blue}{September}} (1.4\text{e-4}) <s_3> & \notag \\ 
    \text{If I want to visit Paris in \textcolor{brown}{spring}, I would like go in} & \notag \\ 
    \text{\textcolor{red}{September}} (9.9\text{e-6}) <s_5> 
\end{align} 
In this case, although the phase of `September' becomes inconsistent when the context phase switches from `autumn' to `spring'. Correspondingly, its probability decreases and its skepticism token levels up, indicating {\ModelName} successfully captures the skepticism implied by the expressions.

\section{Related Work}
\label{related_work}

\subsection{Hallucination Detection}


Many LLM-based studies for hallucination detection are based on internal states \citep{Azaria2023The,huang2023survey,ling2024uncertaintyquantificationincontextlearning,liu2024uncertaintyestimationquantificationllms, su-etal-2024-unsupervised}. By analyzing the minimal token probability within key concepts, \citep{varshney2023stitch} assess the uncertainty of the model towards these concepts. Our {\ModelName} also use token probability, however, we combine both token probability and token information to estimate uncertainty. 



Finetuning LLMs can be useful for uncertainty estimation. Lin et al. train LLM to directly output verbalized probability with CalibratedMath, which is a suite of elementary mathematics problems. LLM’s empirical accuracy on each type of question was used as the label \citep{Lin2022Teaching}. However, their method does not use token probability information. On the other hand, we combine both token probability and token information to finetune the LLM. Kadavath et al. add an auxiliary value head to the LLMs and finetune the models to predict the probability that they can answer a question correctly \citep{kadavath2022languagemodelsmostlyknow}. Nevertheless, they only use questions to train the model, while we use both question and answer.

\subsection{Hallucination Mitigation}

Many methods have been proposed for hallucination mitigation in LLM \citep{Ji2023Towards,dhuliawala-etal-2024-chain,zhang-etal-2024-truthx,zhang-etal-2024-self} . No matter whether LLMs know the knowledge or not, traditional fine-tuning approaches force LLMs to complete a sentence. If the question is beyond the inherent knowledge of LLMs, LLMs will try to fabricate plausible-sounding but mistaken facts. Motivated by this, Zhang et al. propose a method called Refusal-Aware Instruction Tuning (R-Tuning), which constructs a refusal-aware dataset by comparing the prediction and label, and then finetune LLMs to admit their uncertainty about the answer or refuse questions beyond its internal knowledge \citep{Zhang2024R-Tuning}. Elaraby et al. explore teacher-student and knowledge injection methods to mitigate hallucinations in LLMs \citep{Elaraby2023Halo:}. Guan et al. present Knowledge Graph-based Retrofitting (KGR), an approach that use knowledge graph to retrofit the initial responses of LLMs \citep{Guan2024Mitigating}. RL finetuning can also mitigate hallucination \citep{Roit2023Factually,Sun2023Aligning}, However, when facing unfamiliar inputs, reward models may suffer from hallucinations. To tackle this challenge, Kang et al. propose a conservative reward model approach to avoid overestimating rewards for unfamiliar inputs, then use this approach to teach LLMs to generate reliable long-form responses on long text generation tasks \citep{Kang2024Unfamiliar}. 

Comparing to these methods, our {\ModelName} is a pure model-based approach which does not need an external knowledge base. Similar to R-Tuning, our {\ModelName} also augment the original QA with an extra prompt which further asks about the LLM's confidence. However, {\ModelName} include auto-regressive modeling of skepticism tokens and corresponding pretraining, which further strengthen the method's self-skepticism capability.





\section{Conclusion}
\label{conclusion}

In this paper, we introduced a novel self-skepticism and self-aware method for large language models (LLMs) named {\ModelName}. To achieve the skeptical thinking of LLM, we integrate the skepticism tokens into the tokenizer vocabulary, and adapt the LLM to learn to decode both normal tokens and skepticism tokens. Both CPT and SFT stages are conducted, which empowers LLMs to acknowledge their epistemic boundaries by responding with "unsure" when faced with questions beyond their knowledge boundary. This not only mitigates the risk of LLM hallucination but also fosters a more reliable interaction pattern with human users. Through extensive quantitative analysis, we demonstrated the superiority of our method across various data formats, domains and tasks, compared to vanilla fine-tuning method and R-tuning.

\newpage

\bibliography{iclr2025_conference}

\begin{thebibliography}{42}
\providecommand{\natexlab}[1]{#1}
\providecommand{\url}[1]{\texttt{#1}}
\expandafter\ifx\csname urlstyle\endcsname\relax
  \providecommand{\doi}[1]{doi: #1}\else
  \providecommand{\doi}{doi: \begingroup \urlstyle{rm}\Url}\fi

\bibitem[Azaria \& Mitchell(2023)Azaria and Mitchell]{Azaria2023The}
Amos Azaria and Tom Mitchell.
\newblock The internal state of an llm knows when it's lying.
\newblock \emph{EMNLP 2023}, Findings of the Association for Computational Linguistics: EMNLP 2023:\penalty0 967–976, 2023.

\bibitem[Bai et~al.(2024)Bai, Wang, Xiao, He, Han, Zhang, and Shou]{bai2024hallucinationmultimodallargelanguage}
Zechen Bai, Pichao Wang, Tianjun Xiao, Tong He, Zongbo Han, Zheng Zhang, and Mike~Zheng Shou.
\newblock Hallucination of multimodal large language models: A survey, 2024.
\newblock URL \url{https://arxiv.org/abs/2404.18930}.

\bibitem[Cohen et~al.(2023)Cohen, Hamri, Geva, and Globerson]{cohen-etal-2023-lm}
Roi Cohen, May Hamri, Mor Geva, and Amir Globerson.
\newblock {LM} vs {LM}: Detecting factual errors via cross examination.
\newblock In Houda Bouamor, Juan Pino, and Kalika Bali (eds.), \emph{Proceedings of the 2023 Conference on Empirical Methods in Natural Language Processing}, pp.\  12621--12640, Singapore, December 2023. Association for Computational Linguistics.
\newblock \doi{10.18653/v1/2023.emnlp-main.778}.
\newblock URL \url{https://aclanthology.org/2023.emnlp-main.778/}.

\bibitem[Descartes(1641)]{descartes1641meditations}
Rene Descartes.
\newblock Meditations on first philosophy.
\newblock 1641.

\bibitem[Dhuliawala et~al.(2024)Dhuliawala, Komeili, Xu, Raileanu, Li, Celikyilmaz, and Weston]{dhuliawala-etal-2024-chain}
Shehzaad Dhuliawala, Mojtaba Komeili, Jing Xu, Roberta Raileanu, Xian Li, Asli Celikyilmaz, and Jason Weston.
\newblock Chain-of-verification reduces hallucination in large language models.
\newblock In Lun-Wei Ku, Andre Martins, and Vivek Srikumar (eds.), \emph{Findings of the Association for Computational Linguistics: ACL 2024}, pp.\  3563--3578, Bangkok, Thailand, August 2024. Association for Computational Linguistics.
\newblock \doi{10.18653/v1/2024.findings-acl.212}.
\newblock URL \url{https://aclanthology.org/2024.findings-acl.212}.

\bibitem[Elaraby et~al.(2023)Elaraby, Lu, Dunn, Zhang, Wang, Liu, Tian, Wang, and Wang]{Elaraby2023Halo:}
Mohamed Elaraby, Mengyin Lu, Jacob Dunn, Xueying Zhang, Yu~Wang, Shizhu Liu, Pingchuan Tian, Yuping Wang, and Yuxuan Wang.
\newblock Halo: Estimation and reduction of hallucinations in open-source weak large language models.
\newblock \emph{CoRR}, abs/2308.11764, 2023.

\bibitem[Elazar et~al.(2021)Elazar, Kassner, Ravfogel, Ravichander, Hovy, Schutze, and Goldberg]{Elazar2021Measuring}
Yanai Elazar, Nora Kassner, Shauli Ravfogel, Abhilasha Ravichander, Eduard Hovy, Hinrich Schutze, and Yoav Goldberg.
\newblock Measuring and improving consistency in pretrained language models.
\newblock \emph{Transactions of the Association for Computational Linguistics}, 9, 2021.

\bibitem[et~al(2018)]{Thorne2018FEVER:}
James~Thorne et~al.
\newblock Fever: a large-scale dataset for fact extraction and verification.
\newblock \emph{Proceedings of the 2018 Conference of the North American Chapter of the Association for Computational Linguistics Human Language Technologies, Volume 1 (Long Papers)}, 2018.

\bibitem[et~al(2023)]{Kamoi2023WiCE:}
Ryo~Kamoi et~al.
\newblock Wice: Real-world entailment for claims in wikipedia.
\newblock \emph{Computing Research Repository}, 2023.

\bibitem[Facione(1990)]{facione1990critical}
Peter~A Facione.
\newblock Critical thinking: A statement of expert consensus for purposes of educational assessment and instruction. research findings and recommendations.
\newblock 1990.

\bibitem[Feynman(1974)]{Feynman1974notfool}
Richard Feynman.
\newblock Cargo cult science.
\newblock In \emph{Caltech commencement address}, Caltech, California, 1974.

\bibitem[Gao et~al.(2020)Gao, Biderman, Black, Golding, Hoppe, Foster, Phang, He, Thite, Nabeshima, Presser, and Leahy]{Gao2020The}
Leo Gao, Stella Biderman, Sid Black, Laurence Golding, Travis Hoppe, Charles Foster, Jason Phang, Horace He, Anish Thite, Noa Nabeshima, Shawn Presser, and Connor Leahy.
\newblock The pile: An 800gb dataset of diverse text for language modeling.
\newblock \emph{CoRR}, abs/2101.00027, 2020.

\bibitem[Guan et~al.(2024)Guan, Liu, Lin, Lu, He, Han, and Sun]{Guan2024Mitigating}
Xinyan Guan, Yanjiang Liu, Hongyu Lin, Yaojie Lu, Ben He, Xianpei Han, and Le~Sun.
\newblock Mitigating large language model hallucinations via autonomous knowledge graph-based retrofitting.
\newblock \emph{THIRTY-EIGHTH AAAI CONFERENCE ON ARTIFICIAL INTELLIGENCE, VOL 38 NO 16}, 2024.

\bibitem[Hendrycks et~al.(2020)Hendrycks, Burns, Basart, Zou, Mazeika, Song, and Steinhardt]{Hendrycks2020Measuring}
Dan Hendrycks, Collin Burns, Steven Basart, Andy Zou, Mantas Mazeika, Dawn Song, and Jacob Steinhardt.
\newblock Measuring massive multitask language understanding.
\newblock In \emph{International Conference on Learning Representations}, 2020.

\bibitem[Huang et~al.(2023{\natexlab{a}})Huang, Yu, Ma, Zhong, Feng, Wang, Chen, Peng, Feng, Qin, and Liu]{huang2023survey}
Lei Huang, Weijiang Yu, Weitao Ma, Weihong Zhong, Zhangyin Feng, Haotian Wang, Qianglong Chen, Weihua Peng, Xiaocheng Feng, Bing Qin, and Ting Liu.
\newblock A survey on hallucination in large language models: Principles, taxonomy, challenges, and open questions.
\newblock \emph{CoRR}, abs/2311.05232, 2023{\natexlab{a}}.

\bibitem[Huang et~al.(2023{\natexlab{b}})Huang, Yu, Ma, Zhong, Feng, Wang, Chen, Peng, Feng, Qin, and Liu]{huang2023surveyhallucinationlargelanguage}
Lei Huang, Weijiang Yu, Weitao Ma, Weihong Zhong, Zhangyin Feng, Haotian Wang, Qianglong Chen, Weihua Peng, Xiaocheng Feng, Bing Qin, and Ting Liu.
\newblock A survey on hallucination in large language models: Principles, taxonomy, challenges, and open questions, 2023{\natexlab{b}}.
\newblock URL \url{https://arxiv.org/abs/2311.05232}.

\bibitem[Ji et~al.(2023{\natexlab{a}})Ji, Lee, Frieske, Yu, Su, Xu, Ishii, Bang, Madotto, and Fung]{Ji2023Survey}
Ziwei Ji, Nayeon Lee, Rita Frieske, Tiezheng Yu, Dan Su, Yan Xu, Etsuko Ishii, Ye~Jin Bang, Andrea Madotto, and Pascale Fung.
\newblock Survey of hallucination in natural language generation.
\newblock \emph{ACM Computing Surveys}, 55\penalty0 (12):\penalty0 1--38, 2023{\natexlab{a}}.

\bibitem[Ji et~al.(2023{\natexlab{b}})Ji, Yu, Xu, Lee, Ishii, and Fung]{Ji2023Towards}
Ziwei Ji, Tiezheng Yu, Yan Xu, Nayeon Lee, Etsuko Ishii, and Pascale Fung.
\newblock Towards mitigating llm hallucination via self reflection.
\newblock 2023{\natexlab{b}}.

\bibitem[Kadavath et~al.(2022)Kadavath, Conerly, Askell, Henighan, Drain, Perez, Schiefer, Hatfield-Dodds, DasSarma, Tran-Johnson, Johnston, El-Showk, Jones, Elhage, Hume, Chen, Bai, Bowman, Fort, Ganguli, Hernandez, Jacobson, Kernion, Kravec, Lovitt, Ndousse, Olsson, Ringer, Amodei, Brown, Clark, Joseph, Mann, McCandlish, Olah, and Kaplan]{kadavath2022languagemodelsmostlyknow}
Saurav Kadavath, Tom Conerly, Amanda Askell, Tom Henighan, Dawn Drain, Ethan Perez, Nicholas Schiefer, Zac Hatfield-Dodds, Nova DasSarma, Eli Tran-Johnson, Scott Johnston, Sheer El-Showk, Andy Jones, Nelson Elhage, Tristan Hume, Anna Chen, Yuntao Bai, Sam Bowman, Stanislav Fort, Deep Ganguli, Danny Hernandez, Josh Jacobson, Jackson Kernion, Shauna Kravec, Liane Lovitt, Kamal Ndousse, Catherine Olsson, Sam Ringer, Dario Amodei, Tom Brown, Jack Clark, Nicholas Joseph, Ben Mann, Sam McCandlish, Chris Olah, and Jared Kaplan.
\newblock Language models (mostly) know what they know, 2022.
\newblock URL \url{https://arxiv.org/abs/2207.05221}.

\bibitem[Kang et~al.(2024)Kang, Wallace, Tomlin, Kumar, and Levine]{Kang2024Unfamiliar}
Katie Kang, Eric Wallace, Claire Tomlin, Aviral Kumar, and Sergey Levine.
\newblock Unfamiliar finetuning examples control how language models hallucinate.
\newblock \emph{CoRR}, abs/2403.05612, 2024.

\bibitem[Koriat \& Levy-Sadot(1999)Koriat and Levy-Sadot]{koriat1999information}
Asher Koriat and Ravit Levy-Sadot.
\newblock Information-based and experience-based monitoring of one’s own knowledge.
\newblock \emph{Dual-process theories in social psychology}, pp.\  483--502, 1999.

\bibitem[Lin et~al.(2022)Lin, Hilton, and Evans]{Lin2022Teaching}
Stephanie Lin, Jacob Hilton, and Owain Evans.
\newblock Teaching models to express their uncertainty in words.
\newblock \emph{Trans. Mach. Learn. Res.}, 2022, 2022.

\bibitem[Ling et~al.(2024)Ling, Zhao, Zhang, Cheng, Liu, Sun, Oishi, Osaki, Matsuda, Ji, Bai, Zhao, and Chen]{ling2024uncertaintyquantificationincontextlearning}
Chen Ling, Xujiang Zhao, Xuchao Zhang, Wei Cheng, Yanchi Liu, Yiyou Sun, Mika Oishi, Takao Osaki, Katsushi Matsuda, Jie Ji, Guangji Bai, Liang Zhao, and Haifeng Chen.
\newblock Uncertainty quantification for in-context learning of large language models, 2024.
\newblock URL \url{https://arxiv.org/abs/2402.10189}.

\bibitem[Liu et~al.(2024)Liu, Pan, Li, and Chen]{liu2024uncertaintyestimationquantificationllms}
Linyu Liu, Yu~Pan, Xiaocheng Li, and Guanting Chen.
\newblock Uncertainty estimation and quantification for llms: A simple supervised approach, 2024.
\newblock URL \url{https://arxiv.org/abs/2404.15993}.

\bibitem[Minaee et~al.(2024)Minaee, Mikolov, Nikzad, Chenaghlu, Socher, Amatriain, and Gao]{minaee2024largelanguagemodelssurvey}
Shervin Minaee, Tomas Mikolov, Narjes Nikzad, Meysam Chenaghlu, Richard Socher, Xavier Amatriain, and Jianfeng Gao.
\newblock Large language models: A survey, 2024.
\newblock URL \url{https://arxiv.org/abs/2402.06196}.

\bibitem[Naveed et~al.(2024)Naveed, Khan, Qiu, Saqib, Anwar, Usman, Akhtar, Barnes, and Mian]{naveed2024comprehensiveoverviewlargelanguage}
Humza Naveed, Asad~Ullah Khan, Shi Qiu, Muhammad Saqib, Saeed Anwar, Muhammad Usman, Naveed Akhtar, Nick Barnes, and Ajmal Mian.
\newblock A comprehensive overview of large language models, 2024.
\newblock URL \url{https://arxiv.org/abs/2307.06435}.

\bibitem[OpenAI~Team(2024)]{openai2024gpt4technicalreport}
OpenAI OpenAI~Team.
\newblock Gpt-4 technical report, 2024.
\newblock URL \url{https://arxiv.org/abs/2303.08774}.

\bibitem[Peng et~al.(2023)Peng, Galley, He, Cheng, Xie, Hu, Huang, Liden, Yu, Chen, and Gao]{peng2023checkfactstryagain}
Baolin Peng, Michel Galley, Pengcheng He, Hao Cheng, Yujia Xie, Yu~Hu, Qiuyuan Huang, Lars Liden, Zhou Yu, Weizhu Chen, and Jianfeng Gao.
\newblock Check your facts and try again: Improving large language models with external knowledge and automated feedback, 2023.
\newblock URL \url{https://arxiv.org/abs/2302.12813}.

\bibitem[Qwen~Team(2024)]{qwen2techreport2023}
Alibaba~Group Qwen~Team.
\newblock {QWEN2 TECHNICAL REPORT}.
\newblock Technical report, Alibaba Group, 2024.

\bibitem[Roit et~al.(2023)Roit, Ferret, Shani, Aharoni, Cideron, Dadashi, Geist, Girgin, Hussenot, Keller, Momchev, Ramos, Stanczyk, Vieillard, Bachem, Elidan, Hassidim, Pietquin, and Szpektor]{Roit2023Factually}
Paul Roit, Johan Ferret, Lior Shani, Roee Aharoni, Geoffrey Cideron, Robert Dadashi, Matthieu Geist, Sertan Girgin, Léonard Hussenot, Orgad Keller, Nikola Momchev, Sabela Ramos, Piotr Stanczyk, Nino Vieillard, Olivier Bachem, Gal Elidan, Avinatan Hassidim, Olivier Pietquin, and Idan Szpektor.
\newblock Factually consistent summarization via reinforcement learning with textual entailment feedback.
\newblock \emph{Computing Research Repository}, abs/2306.00186, 2023.

\bibitem[Schwarz \& Clore(1983)Schwarz and Clore]{schwarz1983mood}
Norbert Schwarz and Gerald~L Clore.
\newblock Mood, misattribution, and judgments of well-being: Informative and directive functions of affective states.
\newblock \emph{Journal of personality and social psychology}, 45\penalty0 (3):\penalty0 513, 1983.

\bibitem[Soldaini et~al.(2024)Soldaini, Kinney, Bhagia, Schwenk, Atkinson, Authur, Bogin, Chandu, Dumas, Elazar, Hofmann, Jha, Kumar, Lucy, Lyu, Lambert, Magnusson, Morrison, Muennighoff, Naik, Nam, Peters, Ravichander, Richardson, Shen, Strubell, Subramani, Tafjord, Walsh, Zettlemoyer, Smith, Hajishirzi, Beltagy, Groeneveld, Dodge, and Lo]{soldaini2024dolmaopencorpustrillion}
Luca Soldaini, Rodney Kinney, Akshita Bhagia, Dustin Schwenk, David Atkinson, Russell Authur, Ben Bogin, Khyathi Chandu, Jennifer Dumas, Yanai Elazar, Valentin Hofmann, Ananya~Harsh Jha, Sachin Kumar, Li~Lucy, Xinxi Lyu, Nathan Lambert, Ian Magnusson, Jacob Morrison, Niklas Muennighoff, Aakanksha Naik, Crystal Nam, Matthew~E. Peters, Abhilasha Ravichander, Kyle Richardson, Zejiang Shen, Emma Strubell, Nishant Subramani, Oyvind Tafjord, Pete Walsh, Luke Zettlemoyer, Noah~A. Smith, Hannaneh Hajishirzi, Iz~Beltagy, Dirk Groeneveld, Jesse Dodge, and Kyle Lo.
\newblock Dolma: an open corpus of three trillion tokens for language model pretraining research, 2024.
\newblock URL \url{https://arxiv.org/abs/2402.00159}.

\bibitem[Su et~al.(2024)Su, Wang, Ai, Hu, Wu, Zhou, and Liu]{su-etal-2024-unsupervised}
Weihang Su, Changyue Wang, Qingyao Ai, Yiran Hu, Zhijing Wu, Yujia Zhou, and Yiqun Liu.
\newblock Unsupervised real-time hallucination detection based on the internal states of large language models.
\newblock In Lun-Wei Ku, Andre Martins, and Vivek Srikumar (eds.), \emph{Findings of the Association for Computational Linguistics: ACL 2024}, pp.\  14379--14391, Bangkok, Thailand, August 2024. Association for Computational Linguistics.
\newblock \doi{10.18653/v1/2024.findings-acl.854}.
\newblock URL \url{https://aclanthology.org/2024.findings-acl.854}.

\bibitem[Sun et~al.(2023)Sun, Shen, Cao, Liu, Li, Shen, Gan, Gui, Wang, Yang, Keutzer, and Darrell]{Sun2023Aligning}
Zhiqing Sun, Sheng Shen, Shengcao Cao, Haotian Liu, Chunyuan Li, Yikang Shen, Chuang Gan, Liang-Yan Gui, Yu-Xiong Wang, Yiming Yang, Kurt Keutzer, and Trevor Darrell.
\newblock Aligning large multimodal models with factually augmented rlhf.
\newblock \emph{arXiv (Cornell University)}, abs/2309.14525:\penalty0 13088--13110, 2023.

\bibitem[Touvron et~al.(2023)Touvron, Lavril, Izacard, Martinet, Lachaux, Lacroix, Rozière, Goyal, Hambro, Azhar, Rodriguez, Joulin, Grave, and Lample]{touvron2023llamaopenefficientfoundation}
Hugo Touvron, Thibaut Lavril, Gautier Izacard, Xavier Martinet, Marie-Anne Lachaux, Timothée Lacroix, Baptiste Rozière, Naman Goyal, Eric Hambro, Faisal Azhar, Aurelien Rodriguez, Armand Joulin, Edouard Grave, and Guillaume Lample.
\newblock Llama: Open and efficient foundation language models, 2023.
\newblock URL \url{https://arxiv.org/abs/2302.13971}.

\bibitem[Varshney et~al.(2023)Varshney, Yao, Zhang, Chen, and Yu]{varshney2023stitch}
Neeraj Varshney, Wenlin Yao, Hongming Zhang, Jianshu Chen, and Dong Yu.
\newblock A stitch in time saves nine: Detecting and mitigating hallucinations of llms by validating low-confidence generation, 2023.

\bibitem[Yang et~al.(2018)Yang, Peng, Zhang, Bengiov, Cohent, Salakhutdinov, and Manning]{Yang2018HOTPOTQA:}
Zhilin Yang, Qi~Peng, Saizheng Zhang, Yoshua Bengiov, William~W. Cohent, Ruslan Salakhutdinov, and Christopher~D. Manning.
\newblock Hotpotqa: A dataset for diverse, explainable multi-hop question answering.
\newblock \emph{Computing Research Repository}, pp.\  2369--2380, 2018.

\bibitem[Zhang et~al.(2024{\natexlab{a}})Zhang, Diao, Lin, Fung, Lian, Wang, Chen, Ji, and Zhang]{Zhang2024R-Tuning}
Hanning Zhang, Shizhe Diao, Yong Lin, Yi~R. Fung, Qing Lian, Xingyao Wang, Yangyi Chen, Heng Ji, and Tong Zhang.
\newblock R-tuning: Instructing large language models to say `i don't know'.
\newblock \emph{NAACL-HLT}, pp.\  7113--7139, 2024{\natexlab{a}}.

\bibitem[Zhang et~al.(2024{\natexlab{b}})Zhang, Yu, and Feng]{zhang-etal-2024-truthx}
Shaolei Zhang, Tian Yu, and Yang Feng.
\newblock {T}ruth{X}: Alleviating hallucinations by editing large language models in truthful space.
\newblock In Lun-Wei Ku, Andre Martins, and Vivek Srikumar (eds.), \emph{Proceedings of the 62nd Annual Meeting of the Association for Computational Linguistics (Volume 1: Long Papers)}, pp.\  8908--8949, Bangkok, Thailand, August 2024{\natexlab{b}}. Association for Computational Linguistics.
\newblock \doi{10.18653/v1/2024.acl-long.483}.
\newblock URL \url{https://aclanthology.org/2024.acl-long.483}.

\bibitem[Zhang et~al.(2024{\natexlab{c}})Zhang, Peng, Tian, Zhou, Jin, Song, Mi, and Meng]{zhang-etal-2024-self}
Xiaoying Zhang, Baolin Peng, Ye~Tian, Jingyan Zhou, Lifeng Jin, Linfeng Song, Haitao Mi, and Helen Meng.
\newblock Self-alignment for factuality: Mitigating hallucinations in {LLM}s via self-evaluation.
\newblock In Lun-Wei Ku, Andre Martins, and Vivek Srikumar (eds.), \emph{Proceedings of the 62nd Annual Meeting of the Association for Computational Linguistics (Volume 1: Long Papers)}, pp.\  1946--1965, Bangkok, Thailand, August 2024{\natexlab{c}}. Association for Computational Linguistics.
\newblock \doi{10.18653/v1/2024.acl-long.107}.
\newblock URL \url{https://aclanthology.org/2024.acl-long.107}.

\bibitem[Zhang et~al.(2023)Zhang, Li, Cui, Cai, Liu, Fu, Huang, Zhao, Zhang, Chen, Wang, Luu, Bi, Shi, and Shi]{Zhang2023Siren's}
Yue Zhang, Yafu Li, Leyang Cui, Deng Cai, Lemao Liu, Tingchen Fu, Xinting Huang, Enbo Zhao, Yu~Zhang, Yulong Chen, Longyue Wang, Anh~Tuan Luu, Wei Bi, Freda Shi, and Shuming Shi.
\newblock Siren's song in the ai ocean: A survey on hallucination in large language models.
\newblock \emph{CoRR}, abs/2309.01219, 2023.

\bibitem[Zhao et~al.(2023)Zhao, Zhou, Li, Tang, Wang, Hou, Min, Zhang, Zhang, Dong, Du, Yang, Chen, Chen, Jiang, Ren, Li, Tang, Liu, Liu, Nie, and Wen]{zhao2023surveylargelanguagemodels}
Wayne~Xin Zhao, Kun Zhou, Junyi Li, Tianyi Tang, Xiaolei Wang, Yupeng Hou, Yingqian Min, Beichen Zhang, Junjie Zhang, Zican Dong, Yifan Du, Chen Yang, Yushuo Chen, Zhipeng Chen, Jinhao Jiang, Ruiyang Ren, Yifan Li, Xinyu Tang, Zikang Liu, Peiyu Liu, Jian-Yun Nie, and Ji-Rong Wen.
\newblock A survey of large language models, 2023.
\newblock URL \url{https://arxiv.org/abs/2303.18223}.

\end{thebibliography}
\bibliographystyle{iclr2025_conference}

\newpage

\appendix

\section{Additional Implementation Details}

Table \ref{hyperparameter} lists the hyperparameters of experiments. The training epoch is set to 1 and the temperature is $0$. The skeptical threshold $\epsilon$ is set to 0.5.

\begin{table}[t]
\caption{Hyper-parameters of experiments.}
\label{hyperparameter}

\centering
\small
\begin{tabular}{c|c|c}
\toprule
Stage & Parameters & Value \\ 
\toprule
        
\multirow{3}{*}{CPT} & learning rate & 5e-7 \\
   & weight decay & 0.01 \\
   & batch size & 1024 \\
   
\cmidrule{1-3}
\multirow{3}{*}{SFT} & learning rate & 1e-6 \\
  & weight decay & 0.01 \\
    & batch size & 128 \\
\bottomrule
\end{tabular}
\end{table}

\section{Additional Results}

\begin{figure*}[htbp!]
    \centering
    \includegraphics[width=0.48\linewidth]{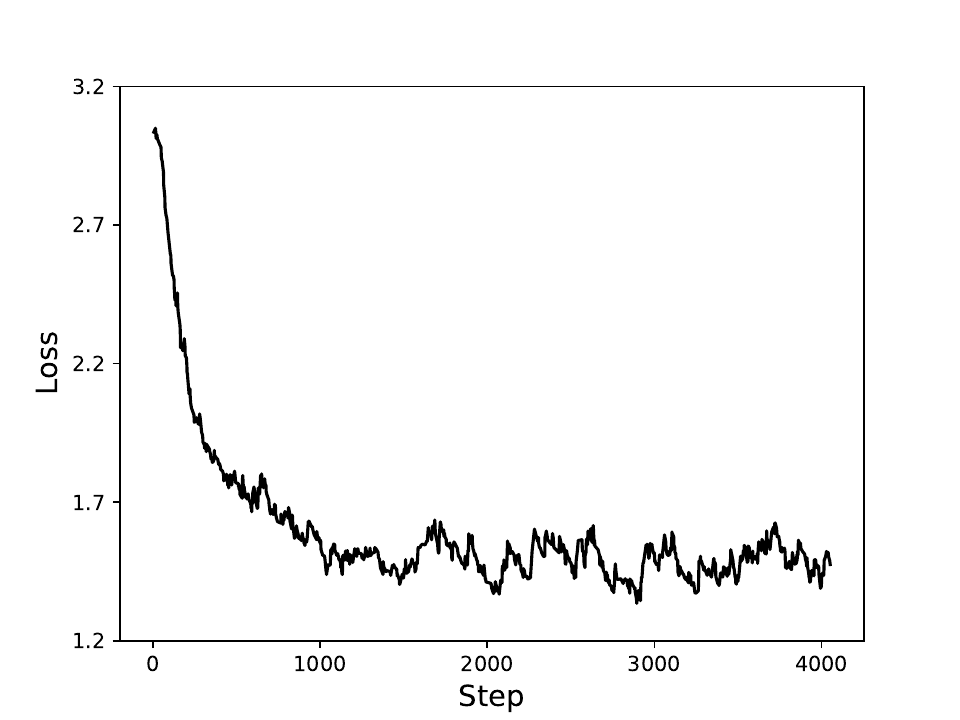}
    \hspace{0.1in}
    \includegraphics[width=0.48\linewidth]{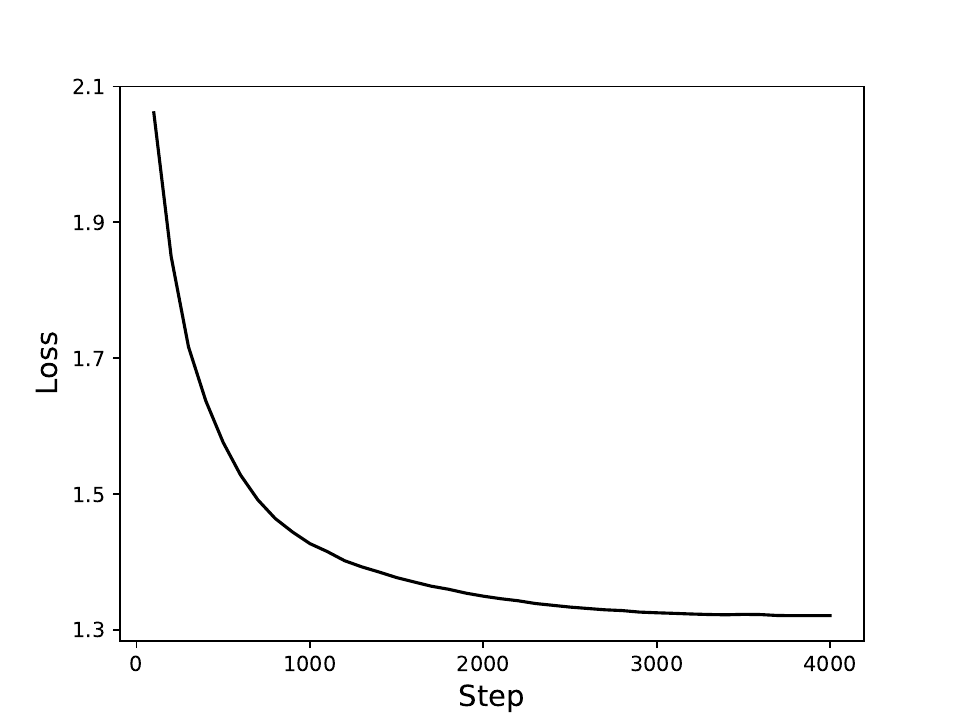}
    \caption{Loss Curves of the CPT stage of {\ModelName}. Left: the training set; Right: the test set.}
    \label{fig:losses}
\end{figure*}

Figure \ref{fig:losses} shows the training and evaluation losses during the CPT stage. {\ModelName} successfully converges with the new skepticism tokens added into the vacabulary.

\begin{figure*}[h!]
    \centering
    \includegraphics[width=0.485\linewidth]{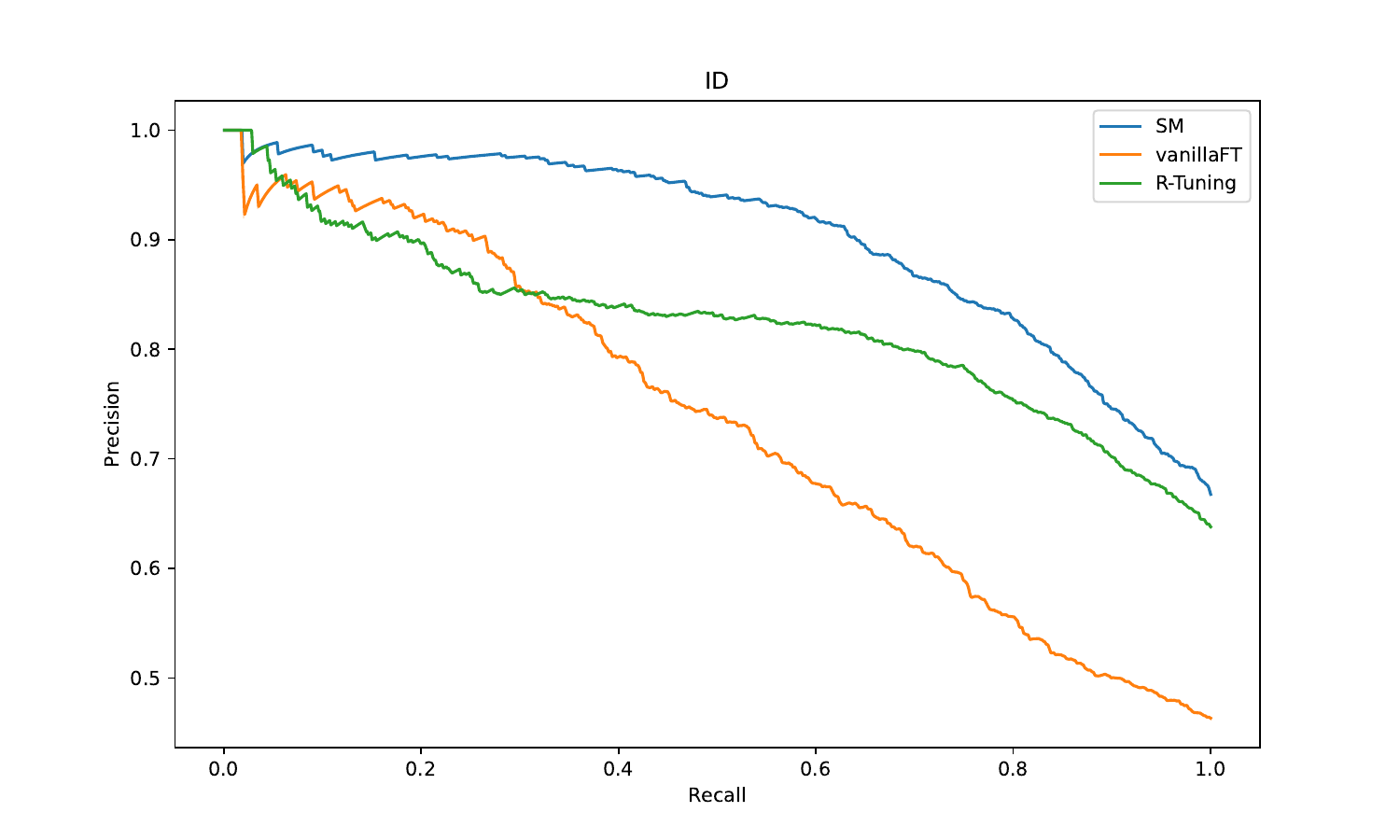}
    \hspace{0.1in}
    \includegraphics[width=0.485\linewidth]{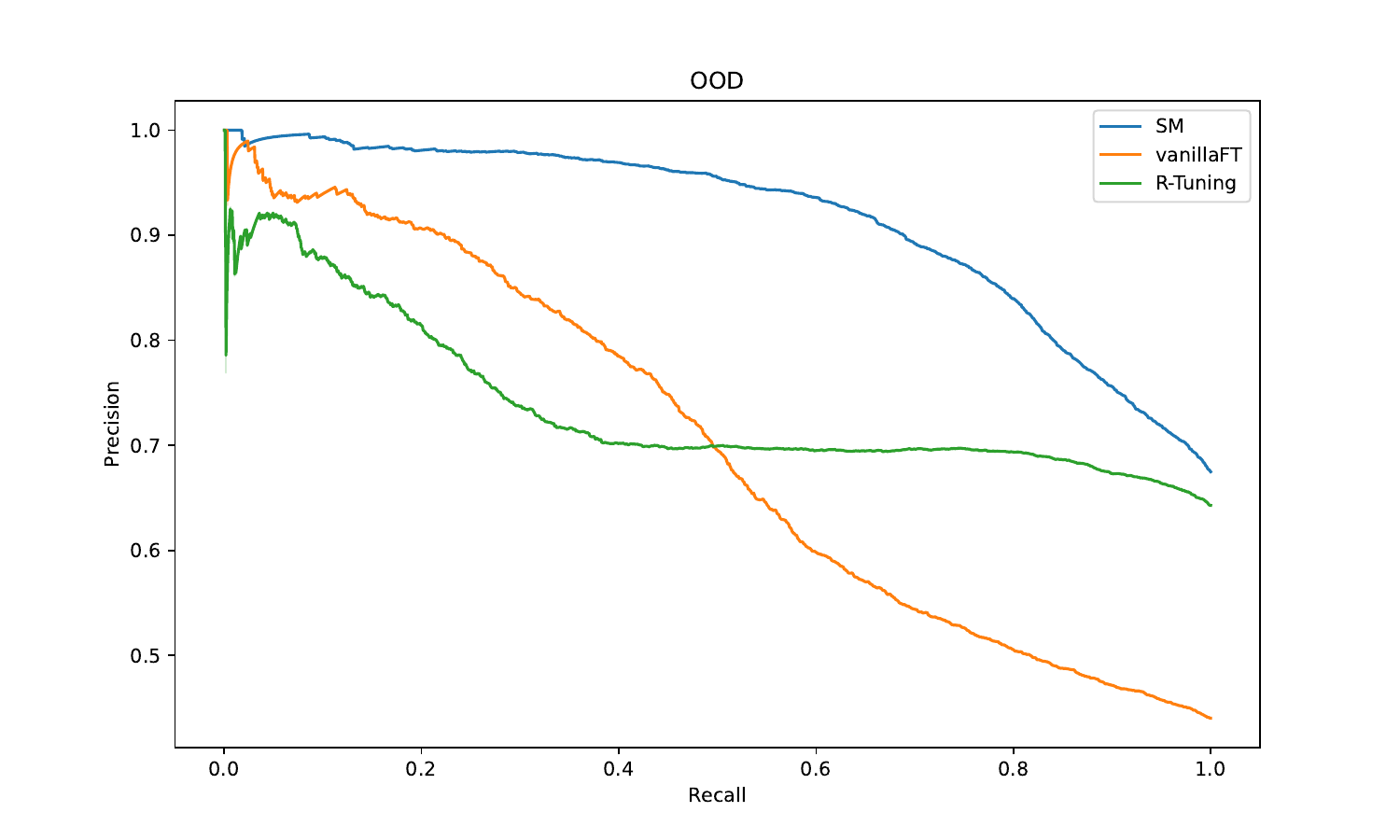}
    \caption{Multitask Experimental Precision-Recall Curves on MMLU, with ID and OOD Subsets.}
    \label{multi-task}
\end{figure*}

We also conduct multi-task experiments and exhibit the Precision-Recall curves on MMLU, with ID and OOD domains, respectively. As indicated by Figure \ref{multi-task},  a higher AP score means better performance. This result indicates our model perform well in multi-task setting and show good generalization ability.

\end{document}